# Hierarchical Random Walker Segmentation for Large Volumetric Biomedical Images

Dominik Drees, Florian Eilers and Xiaoyi Jiang *Senior Member, IEEE*

*Abstract*—The random walker method for image segmentation is a popular tool for semi-automatic image segmentation, especially in the biomedical field. However, its linear asymptotic run time and memory requirements make application to 3D datasets of increasing sizes impractical. We propose a hierarchical framework that, to the best of our knowledge, is the first attempt to overcome these restrictions for the random walker algorithm and achieves sublinear run time and constant memory complexity. The goal of this framework is – rather than improving the segmentation quality compared to the baseline method – to make interactive segmentation on out-of-core datasets possible. The method is evaluated quantitatively on synthetic data and the CT-ORG dataset where the expected improvements in algorithm run time while maintaining high segmentation quality are confirmed. The incremental (i.e., interaction update) run time is demonstrated to be in seconds on a standard PC even for volumes of hundreds of gigabytes in size. In a small case study the applicability to large real world from current biomedical research is demonstrated. An implementation of the presented method is publicly available in version 5.2 of the widely used volume rendering and processing software Voreen[1].

## I. INTRODUCTION

The random walker method for image segmentation [1] remains a hugely popular technique [2], [3] for interactively creating arbitrary segmentations of 3D (or 2D) datasets which inherently supports multi-class segmentation and uncertainty feedback [4]. From user-defined initial seeds for semantically different regions of the image, the complete image is labeled based on local connectivity and voxel-wise similarity. It is frequently used, for example, in the biomedical field [5] where volume segmentation is often an important intermediate step for further analysis (from simple volume/area comparisons to complex structure analysis [6]). At the same time, improvements in imaging technologies continue to provide images of increasing resolution and size. Specifically, light sheet microscopy yields single datasets of hundreds of gigabytes or even terabytes in size. While this opens the door for biomedical research [7], [8], it also poses problems for image analysis algorithms that need to be solved to further advance biomedical and other research with new technology providing quantifiable results.

D. Drees, F. Eilers and X. Jiang are with the Faculty of Mathematics and Computer Science, University of Münster, Münster, Germany.

F. Eilers is a member of CiM-IMPRS, the joint graduate school of the Cells-in-Motion Interfaculty Centre, University of Münster, Germany and the International Max Planck Research School - Molecular Biomedicine, Münster, Germany.

[1]https://www.uni-muenster.de/Voreen/

Specifically, application of the random walker technique on larger datasets is problematic due to its lacking scalability: For a dataset with $n$ voxels, it requires $\Theta(n)$ memory with a high constant factor[2] and $\Omega(n)$ computation time even for a very local update of the set of seeds. In practice, even for datasets smaller than $100\,\text{MB}$, minutes of computation time and large portions of the available graphics or main memory of a standard PC are required, a problem that is not fundamentally solved by earlier extensions to the method [9].

*Contributions:* We present a hierarchical, level-of-detail (LOD) pyramid-based framework that, to the best of our knowledge for the first time, allows random walker image segmentation to be applied to out-of-core (i.e., larger than main memory) datasets. The core idea of hierarchical refinement of an initial coarse application of the basic method in constant sized bricks achieves constant memory requirements while further optimizations such as pruning result in an observed sublinear run time. Using our technique, incremental updates can often be computed in seconds even for $100\,\text{GB}$ and larger volumes on commodity hardware. The goal of this paper is explicitly *not* the improvement of segmentation quality. However, we show that segmentation quality is comparable to the baseline method with our framework. While the description of the method focuses on binary image segmentation, application for multi-class segmentation is also demonstrated. Further, a generalization to 2D data is conceivable, but not considered here.

We believe that the presented method is useful, for example, to apply existing RW-based pipelines [10], [11], [12] to large datasets, or as a highly efficient labeling tool for machine learning applications.

The remainder of the paper is organized as follows. After presenting related work and two fundamental concepts, the proposed framework will be described in detail. Subsequently, an evaluation on synthetic data, the CT-ORG dataset and real data from current biomedical research is presented and discussed, followed by the conclusion.

## II. RELATED WORK

The seminal work of Grady [1] is the basis of a large number of publications with applications in biomedicine (fluorescent cell tracking [13], pap smear cells [14], tumor [15], [16], placenta [12], airway segmentation [10], [17], multimodal problems [18], [19] and geophysical research [20]). It is also used in other algorithms (image superpixel segmentation [21], [22] and deep learning techniques [23]).

[2]Specifically, assuming a conjugate gradient solver (which requires intermediate buffers), Ellpack sparse matrix format (with buffers for values and column indices) solver and using a Jacobi preconditioner (additional matrix and vector buffers) $116n$ bytes are required.

Random walker – as the one of the most popular interactive segmentation methods [3] – has been the basis for numerous extensions: This includes, for example, addition of additional nodes and connections to the image graph for segmentation of elongated structures [24] or for improved computational efficiency [25]. Other research focuses on automatic seed generation [26] or interactive boundary definition [27]. The automatic choice of edge weighting functions has been tackled using statistical models [28], [29] and an end-to-end learning [30]. Li et al. [31] improve the convergence of the random walker method and reduce the seed sensitivity using a continuous random walk model with coherence regularization for the extracted foreground region. However, resource requirements of the random walker method hinder application in some cases: Xiao and Hu [32] apply Grady's method in a two-step ("hierarchical") procedure to first identify an abnormal region and then a tumor within. They note that during the evaluation "12 groups failed due to high space complexity of RW".

In general, hierarchical representations play a role for segmentation in the form of quadtree (2D) or octree (3D) segmentations for volume reconstruction from depth images [33] or point clouds [34], as well as for image compression [35]. In contrast to the proposed method which utilizes a multi-resolution octree structure, these typically aim to *produce* a hierarchical representation for further processing that only store a single intensity per node.

On the other hand, methods operating utilizing a hierarchy of resolutions of the input image have a rich history in image segmentation. Often, scale space filtering [36] is used for capturing of local and global features with otherwise similar methods and has been generalized [37] to scale space construction techniques. As such, it has been applied to the random walker method [38] for noise resilience by combining results from applying the random walker method at different scales, but also for watershed segmentation [39], and more recently in machine learning [40], [41]. Multi-level techniques are also applied to improve the run time of methods via faster label propagation [42] and improved convergence of numerical algorithms [43].

Of special note is the work Lombaert et al. [44] where they propose a hierarchical strategy for the related graph-cuts method [45]. The authors suggest creating an initial segmentation on the coarsest level of a LOD-pyramid and using finer levels for refinement of segmentation boundaries. The goal of the method is the reduction of the graph size, which is only constructed at boundary regions of the coarser segmentation which results in run time savings also reported in derivative works [46]. While this also reduces the memory footprint of the graph, the memory requirements of the overall method and run time complexity of auxiliary operations (e.g., construction of the final image) are not considered further. Delong and Boykov [47] propose a parallel solver for the graph-cuts problem which allows efficient processing of some larger-than-main memory image graphs, although their experiments show that they still hit a "performance wall" if the ratio between data structure size and available main memory becomes too large (larger than $\approx 5$). It should further be noted that graph-cuts-based methods like the above are only applicable for binary image segmentation. A formulation for multi-class segmentation exists [48], but at the cost of exponential memory blow-up [49].

For the random walker method, Fabijanska and Goclawski [50] try to reduce the computational burden and memory requirements by applying it to supervoxels. While reducing the size of the linear equation system by roughly 90%, the preprocessing step still requires random access to the full image and introduces two parameters, changes to which require a complete recomputation, hindering interactive use on large datasets.

Biomedisa [51] is a recent platform for biomedical image segmentation that performs interpolation of sparsely pre-segmented slices based on simulation of random walker agents. Despite testing on a powerful system (750 GB RAM and 4 NVIDIA Tesla V100) the authors had to split a moderately sized dataset of $900 \times 1303 \times 4327$ voxels for processing.

To the best of our knowledge there is no prior work on frameworks that allow the random walker method to be applied to images that exceed the main or graphics memory.

Finally, artificial neural network approaches have become increasingly popular [52], [53] which, depending on the problem, network design and receptive field, may also be suitable for large volume segmentation. Nevertheless, random walker-based methods remain popular [2]. This may be due to the fact that high level of complexity (large amount of ground truth (GT) data, hyperparameters, etc.) of deep learning-based methods hinders fast problem-solving for novel scenarios and diverse image data. Non-learning-based methods like random walker are useful in situations where machine learning is not suitable (few datasets or "ad hoc" segmentation of varying structures of interest), where (at least presently) lacking generalizability hinders application, or where interactive correction of automatic results is required [54]. In other situations, where machine learning is suitable, our work serves as a highly efficient and effective labeling tool.

## III. Fundamental Concepts

This section introduces two known, but vital components of the proposed framework: The octree data structure and the original random walker method.

### A. Octree Data Structure

The problem of high performance rendering of large volumetric datasets is, at least partially, solved in visualization, often using an octree data structure that was originally introduced by LaMar [55] and Weiler [56] and improved in recent years [57], [58], [59]. Volumes are stored on disk as a brick-based LOD-pyramid from which parts can efficiently be loaded into memory for ray casting.

Similarly, the base structure for the LOD-pyramid in this paper is an octree, recursively defined with nodes $n \in N$ where a node has either 8 or 0 children. Voxels at any level of the pyramid are stored in *bricks* of $s^3$ voxels with side length $s$. In the following, no formal distinction is made between a node and its associated brick. We thus define $N = \mathbb{R}^{s^3}$ for use in the context of bricks. Where useful the level of a node within the tree is indicated with a subscript $n_i$ such that $n_0$ is the root of the tree and $n_{i-1}$ always refers to the parent node of $n_i$.

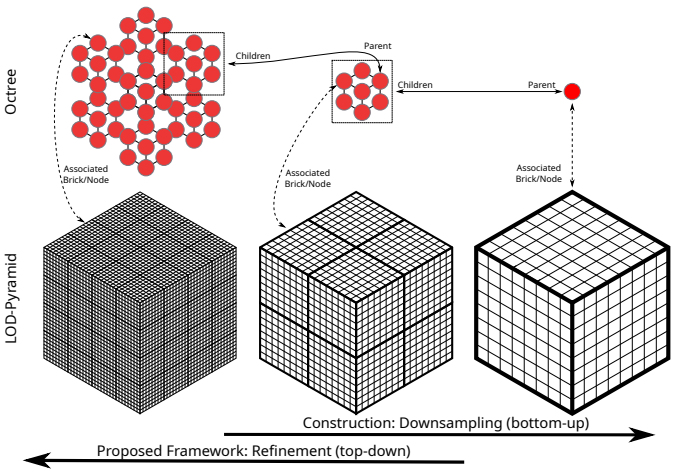

Fig. 1. LOD-Pyramid with octree used for hierarchical processing. The full resolution dataset (bottom left) is divided into bricks of $s^3$ voxels, each brick with an associated node. $2^3$ nodes with neighboring bricks share a parent node. The parent node's brick is computed by downsampling its children's bricks. The root node's brick is a low resolution version (with $s^3$ voxels) of the input dataset (with $64s^3$ voxels in this example). After the bottom-up construction of the octree, our method uses it in a top-down process.

When converting a voxel-grid volume of dimensions $d_x \times d_y \times d_z$ into an octree representation, voxels are chunked into $\frac{d_x}{s} \times \frac{d_y}{s} \times \frac{d_z}{s}$ bricks, each forming a leaf in the octree. The tree is then constructed bottom-up by bundling $2^3$ neighboring nodes of one level as children of a new parent node in the next, coarser level until only a single root node remains. The parent node's associated brick is generated by applying a $2^3$ mean filter and downsampling the combined bricks of its children. Here we assume $d_x = d_y = d_z = 2^i \cdot s$, but other volume dimensions can be handled by dynamic transformation and handling of the volume border in the brick-local space. This data structure is illustrated in Figure 1 and described in more detail by Brix et al. [57]. The construction of the octree representation from a single-resolution dataset can be done in linear time and with sub-linear memory requirements, but is not considered in the following, since this is typically done as a one-time preprocessing step that is also required for rendering, or not required at all in cases where instruments already output multi-resolution files.

### B. Random Walker Volume Segmentation

The random walker image segmentation method [1] can be used to interactively generate arbitrary foreground/background segmentations in 3D (or 2D) images. The volume (or brick) is viewed as a 3D grid of vertices with edges between 6-neighbors. Vertices $p \in \{1, \ldots, s\}^3$ have associated weights defined by the intensity $x(p) = n[p]$, where brackets $n[p]$ denote sampling the intensity from the brick of a node or volume $n$ at position $p$. Edge weights $w_{p,q}$ are based on the similarity of the connected voxel vertices. Grady suggests $w_{p,q} = \exp(-\beta(n[p] - n[q])^2)$ with $\beta$ as a free parameter. When interpreting the image graph as an electrical circuit, edges are resistors with weights defining the conductance between vertices. Labeled vertices $p_m \in V_m$ are voltage sources with a constant voltage of $x(p) \in \{0, 1\}$, which is set to 0 for background labels and 1 for foreground labels. The weight of an unlabeled vertex $p_u \in V_u$, $x(p_u)$, is the resulting voltage and foreground probability. It can be computed by solving the discrete Dirichlet problem, i.e., by minimizing the Dirichlet integral $D(x) = \frac{1}{2} x^T L x = \frac{1}{2} \sum_{||p-q||=1} w_{p,q}(x(p) - x(q))^2$ with the boundary condition of predetermined $x(p_m)$ and the Laplacian matrix $L$ of the image graph. A global minimum can be found by partitioning and reordering $x$ according to $V_u$ and $V_m$ into $x_u$ and $x_m$ and solving $\frac{dD(x_u)}{dx_u} = 0$. While the minimization can be done analytically, it is often faster to use iterative approaches that additionally exploit the sparsity of $L$ [60].

The solution $x$ can also be interpreted as a foreground probability map, which can be thresholded at 0.5 to obtain a class map in the binary segmentation case. For $m$-class segmentation, the procedure is repeated $m$ times with seeds of class $i$ as foreground seeds and the union of all other class seeds as background seeds to obtain the solutions $x_i$. The class of each voxel is obtained using $c = \operatorname*{argmax}_{i \in \{1,\ldots,m\}} x_i$. In other words, the class with the highest probability is assigned.

## IV. HIERARCHICAL FRAMEWORK: METHODS

This section first describes the core concept of our framework that has constant memory requirements, as well as two important extensions of the core idea. Then, a number of further considerations that greatly improve the run time as well as a partial update scheme for interactive segmentation are introduced.

### A. Core Concept

The proposed method processes an input octree top-down and at the same time creates the output octree. In each layer, the random walker method for (3D) image segmentation [1] (or a suitable extension [9]) is used to process a brick of the input layer, thus creating a foreground probability map, one brick at a time. Labels are given as structures such as points or curves. Here, we define individual labels $p \in P = \mathbb{R}^3 \times [0, 1]$ as 3D points with an associated seed value and label sets $l \in L = 2^P$ as groups of labels.

For processing individual bricks, user-defined labels are transformed from volume space into local brick space using a transformation function $\mathcal{T}_n \colon L \to L$ (comprising translation and scaling) defined for each node $n$. After the transformation, labels are rasterized to obtain brick-level seeds to be used in the random walker algorithm.

Let $\mathrm{rw} \colon N \times L \times \mathbb{R} \to N$ be the application of the random walker algorithm to a given input brick using a set of labels (in brick-local coordinates) and parameter $\beta$. Then, in the coarsest layer of the LOD-pyramid, i.e., for the root node, the proposed procedure ($\mathcal{H}\colon N \times L \times \mathbb{R} \to N$) consists simply of the random walker method with transformed labels:

$$\mathcal{H}(n_0, l, \beta) = \mathrm{rw}(n_0, \mathcal{T}_{n_0}(l), \beta) \tag{1}$$

The output is only defined by the input volume, user labels and $\beta$. For subsequent layers, however, the solution for the coarser layer serves as an approximation of the *true* probability map at the current level (if hypothetically computed on the full volume at the current LOD level) and is used to add additional *continuous labels* (with values in the *range* [0,1], but not necessarily either 0 or 1) at the border of the current brick, i.e., by upsampling from the coarser level probability map. This way, *global* context (either directly from the parent layer, or transitively via one of its parents) is introduced into the *local* foreground probability map. This is important because, for example, a brick may contain parts of the surface of a foreground structure without containing a user-provided seed since the structure may be larger than the brick itself. For a specific brick, the set of seeds hence consists of the union of user-provided seeds (either 0 or 1) that lie within that brick and continuous seeds (between 0 and 1) sampled at its border from the parent solution. For non-root nodes $n_i$ we hence define

$$\mathcal{H}(n_i, l, \beta) = \mathrm{rw}(n_i, \mathcal{L}(n_i, l), \beta)) \tag{2}$$

where $\mathcal{L}$ is the combined set of seeds which can be described as $\mathcal{L}(n_i, l) = \mathcal{T}_{n_i}(l) \cup \mathrm{sample\_border}_{n_i}(\mathcal{H}(n_{i-1}, l, \beta))$. An example of how a labeled object influences the probabilities in a neighboring brick is illustrated in Figure 2.

The algorithm finishes after processing all leaf level bricks, where the resulting probability bricks define the foreground probability map of the voxel-grid volume that was used to create the input octree.

The hierarchical, local application of the basic method makes the presented approach applicable to arbitrarily large out-of-core datasets. Specifically, since individual bricks and resulting edge weight matrices are small (depending on $s$), they can be stored in main or graphics memory. At the same time all computations are local to a single brick or its direct neighborhood, thus require few, linear disk accesses and can be GPU-accelerated. We suggest a brick size of $s = 32$ (independent of volume size), which is also used in experiments if not specified otherwise. Larger brick sizes increase the run time and memory requirement, while smaller bricks cause inaccurate results. This is also demonstrated experimentally in subsection V-E.

Multi-class segmentation using our method can be achieved in the same way as Grady suggests for the original method: The method is run once for each class, where only seeds for the current class are considered foreground and all other seeds as background. Again, the class of a voxel is the one that maximizes the foreground probability. Final classes can either be computed in a final computation step over generated octree volumes for all classes, or computed on the fly where required in the octree-based rendering method [58].

It should be noted that the proposed method is not restricted to volumes of size $2^i \cdot s$: During the octree construction, at the border of the volume bricks smaller than $s^3$ voxels and fewer than 8 null references for child nodes are allowed. Similarly, during the top-down segmentation procedure, bricks at the border of the volume may be smaller than $s^3$ voxels. As this technical aspect in not important in the overall description of the method, we assume a cubical volume size that is a power of two and a multiple of $s$ in each dimension.

### B. Neighborhood Expansion

A remaining open question is where the continuous seeds from coarser octree levels should be located. Placing them directly on the border of the current brick means that (since any non-root brick shares 3 of its 6 border faces with its parent, possibly even transitively) in the worst case global context is only established in the root node, i.e. with low resolution. This could result in discontinuities in the resulting probability map. Hence, for the probability map computation the brick is expanded into its neighbors: The expanded brick side length $s_e = \lfloor e \cdot s \rfloor$ is larger than the original brick by a factor $e > 1$. Let $h \in H := \mathbb{R}^{s_e^3}$ be a brick neighborhood, and $\mathcal{N}\colon N \to H$ be the expansion operation which samples values from the neighbors of a brick. Further, let $\mathcal{N}^{-1}\colon H \to N$ be the "inverse" operation which contracts a neighborhood back to the original brick bounds. Concretely, for our method this means for non-root nodes, that we first expand an input brick, apply the basic random walker method using seeds sampled from the parent layer at the border of the neighborhood, and contract the resulting probability map to the $s^3$ sized output brick. We obtain a refined version of Equation 2:

$$\mathcal{H}(n, l, \beta) = \mathcal{N}^{-1}(\mathrm{rw}(\mathcal{N}(n), \mathcal{L}(\mathcal{N}(n, l)), \beta)) \tag{3}$$

A large factor $e$ is advantageous due to the increased high resolution global context, but also results in an increase of computational time by a factor of $\mathcal{O}(e^3)$. We suggest $e = 1.25$ as a balance between accuracy and run time.

### C. Adaptive Parameter Setting

A challenge (but also advantage) of the octree data structure is that noise on the full resolution level is reduced by the mean filtering in the construction of coarser levels. While reduced noise on coarser octree levels is obviously advantageous for application of the random walker method, it also implies that the optimal choice for $\beta$ is dependent on the octree level – and may in fact vary for different regions of the image [29]. Hence, a fully automatic way of computing edge weights is required. This has the additional advantage that weight functions can be adapted to regions with different noise levels within the image. For the experiments in this paper the t-test-based method of Bian et al. [29] was chosen which assumes signal-dependent Gaussian noise and makes this method very versatile. The weight function $w_{p,q}$ is defined as the P-value of the $H_0$ hypothesis that the pixel distributions defined by values sampled around $p$ and $q$ have the same mean. Equation 3 is thus further refined to use automatic weight function $w$ which removes the parameter $\beta$:

$$\mathcal{H}(n, l) = \mathcal{N}^{-1}(\mathrm{rw}_w(\mathcal{N}(n), \mathcal{L}(\mathcal{N}(n, l)))) \tag{4}$$

Other options include another statistical modeling-based weight function [28] which assumes additive Gaussian noise and defines a weight function derived from the first estimated the global variance $\sigma^2$ of pixel intensity differences. Further, in [61] a general noise model dependent framework is proposed. For images with Poisson noise it yields the following weight function which is used in our case study: $w_{p,q} = \exp(-\frac{1}{2}(\sqrt{n[p]} - \sqrt{n[q]})^2)$.

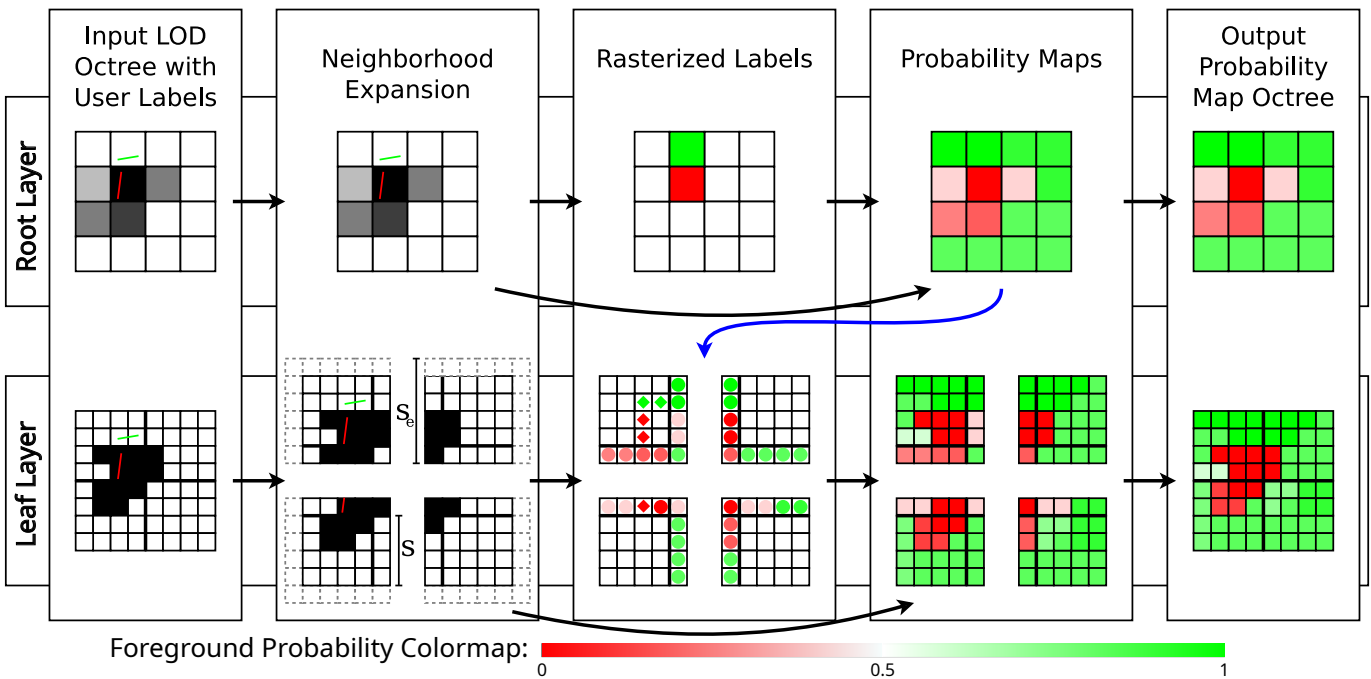

Fig. 2. Exemplary 2D illustration of the proposed method: The input consists of a volume octree (here with black/white voxels in highest resolution and gray level values due to downsampling at coarser scales) and foreground/background labels (green/red lines in this case). Our method then operates by processing the octree top-down, applying the standard random walker method to expanded (but constant-sized) bricks in each layer to obtain a per-brick probability map. Crucially, in addition to user-provided seeds (rasterized from red/green lines and visualized as rhombi), additional continuous seeds at the border of the expanded brick (visualized as disks) are obtained by sampling the solution from the coarser level, thus propagating global context to bricks without user-provided seeds (blue arrow). In a more complex case, intermediate layers would be included in the top-down process, using the probability map of the previous layer and providing global context for the next layer via its probability map.

### D. Run Time Enhancements

While the preliminary framework description above (Equation 4) ensures accuracy and constant memory complexity (since brick are of constant size and are processed one-by-one), this section discusses run time enhancements that directly follow (iterative solver initialization, parallelization) or are enabled by the algorithm structure (pruning).

*1) Pruning:* In almost all cases the label set contains seeds that are targeted at an object that is larger than one brick. Notably, this is also the case when only labeling relatively few, small foreground objects, leaving large *background regions* spanning multiple bricks. In this case it is usually unnecessary to complete the proposed procedure down to the leaf level for all branches of the tree. When the resulting foreground probability of a node $n$ is *homogeneous* (hom: $N \to Bool$), i.e., its range is smaller than a threshold $t_{hom}$, its children do not have to be processed.

$$\text{hom}(n) = (\max(n) - \min(n) < t_{hom})$$

where $\min(n)$ and $\max(n)$ denote the minimum/maximum voxel value in the brick of $n$, respectively. In our experiments the threshold was set to $t_{hom} = 0.3$.

Furthermore, if the user is not interested in the foreground probability volume itself, but in a binary segmentation (by thresholding the probability at $p = 0.5$), nodes do not have to be processed further if the foreground/background state is already determined (dt: $N \to Bool$). This is the case if the lowest foreground probability is above 0.5 (resulting in a brick) or the maximum is below 0.5 (resulting in a foreground brick) by a specific margin $t_{bin}$.

$$\text{dt}(n) = (\min(n) - 0.5 > t_{bin}) \lor (0.5 - \max(n) > t_{bin})$$

For a binary segmentation (i.e., the output probability map thresholded at 0.5) a low value of $t_{bin} = 0.01$ is sufficient.

This pruning process may cause sampling probabilities for continuous seeds at the border of the current brick to fail for neighbors of the parent node since the tree is no longer complete. In this case sampling continues at the nearest transitive parent of the absent node, in the worst case at the root node. Additionally, during the rasterization foreground and background seeds may be in conflict (cfl: $N \times L \to Bool$), i.e., occupy the same voxel:

$$\text{cfl}(n, l) = \exists (q_0, p_0), (q_1, p_1) \in l :$$
$$\lfloor \mathcal{T}_n(q_0) \rfloor = \lfloor \mathcal{T}_n(q_1) \rfloor \land p_0 \neq p_1$$

In this case the voxel is left unseeded and its output probability is determined by other non-conflicting seeds. While this is normally not a problem on the leaf level (i.e., original resolution of the volume) for sensible user-defined seeds, non-conflicting seeds on the finest level can be in conflict when rasterized on a coarser octree level with a lower brick resolution. In the core procedure this is not problematic since the ambiguity is automatically resolved at a finer level. For pruning this means that children of nodes with seed conflicts cannot be pruned and thus have to be processed even if they appear to be homogeneous or determined.

With the above definitions we can define $\mathcal{H}_p$ as an optimized version of $\mathcal{H}$ (from Equation 4):

$$\text{prunable}(n) = (\hom(n) \vee \text{dt}(n)) \wedge \neg \text{cfl}(n,l)$$

$$\mathcal{H}_p(n_i, l) = \begin{cases} \epsilon & \text{prunable}(\mathcal{H}(n_{i-1}, l), l) \\ \mathcal{H}(n_i, l) & \text{else} \end{cases}$$

If foreground and background can clearly be separated in the image (and thus no regions of high uncertainty are generated in the output probability map), the pruning methods described above make sure that only children of bricks that intersect the object surface (i.e., the foreground/background boundary) need to be processed. For a "well behaved" object surface (i.e., with finite surface area, which is of course always the case in a discretized image) the number of bricks that intersect the object boundary is within $\mathcal{O}(k^{\frac{2}{3}})$ if the volume is subdivided into $k$ bricks. A proof of this statement can be found in the appendix. This idea of restricting refinement to the surface of objects has also been applied successfully in the hierarchical graph-cuts method [44]. As the size of a brick is constant and computations performed on the brick are thus within $\mathcal{O}(1)$, the total runtime of the algorithm is within $\mathcal{O}(v^{\frac{2}{3}})$ for a volume of $v = d_x \times d_y \times d_z$ voxels. This behavior is also observed experimentally in section V.

*2) Iterative Solver Initialization:* Even though it is efficient to choose a direct method (e.g. using Cholesky-Decomposition) for computing the random walker method on small 2D images, the more strongly connected 3D lattice reduces the efficiency sufficiently to make it unsuitable even for small brick sizes [62]. A more efficient method for solving sparse systems 3D is (as noted by Grady [1] as well) using an iterative method [60], which can typically be accelerated by providing an initial state that is close to the true solution. For all but the root level this is the case in our method: As described above, for all levels but the root, the foreground probability map computed for the parent brick is available when processing the child, and is usually (barring seed conflicts) a good approximation for the solution of the child. After resampling to the higher resolution of the child brick, it is thus used as an initialization for the iterative solver.

*3) Parallelization:* As all bricks within an octree layer are independent from one another, the computation can trivially be parallelized. While costly brick processing steps (such as filtering, edge weight computation and solving of linear equation systems) are parallelizable themselves, this make the procedure well suited for parallelization on multiple GPUs in workstations.

*E. Incremental Labeling*

The set of user-defined seeds in semi-automatic segmentation algorithms is typically not created in a single step, but built up incrementally in an evaluate-edit-compute cycle, in which the user inspects the existing segmentation, changes the parameters (seeds) of the algorithm and then waits for the algorithm to update the segmentation before starting the cycle again, inspecting the updated segmentation [63]. As such, computation time is not the only factor affecting the total time to produce the desired segmentation result, but still very important, since high latency between user interaction and display of intermediate results impacts user performance [64]. Consequently, the *incremental run time* of an algorithm, in which the previous result is an additional input, has to be considered and evaluated.

Grady [1] suggests using the previous result as initialization for iterative solver for the standard random walker method. This is applicable here (even for non-iterative application, as described above), but not sufficient if the previous result octree is nearly complete: In this case operations such as loading (possibly from disk) still have to be executed for all branches of the tree. Instead, we reuse entire branches of the previous result if they can be expected to be similar to the new result: After computing a new foreground probability map for a node if the result is sufficiently similar to the previous probability map of the corresponding node in the old tree, the new node is discarded and replaced by the old node *including* its children. With $\mathcal{H}_{it}(n, l')$ as the previous solution for label set $l'$ this can be described with the following rule:

$$\text{if } \mathcal{H}(n_i, l) \approx \mathcal{H}_{it}(n_i, l') \wedge \neg \text{cfl}(n_i, l) :$$
$$\forall \text{ children } n_j \text{ of } n_i : \mathcal{H}_{it}(n_j, l) = \mathcal{H}_{it}(n_j, l')$$

Similar to pruning, if there are conflicts in the set of (changed) seeds affecting this brick, this optimization cannot be performed, but it is sufficient to consider conflicts where at least one of the recently added labels $(l - l')$ is involved. Probability maps are considered similar if the maximum pairwise difference is below a threshold $t_{inc}$. We set $t_{inc} = 0.01$ in all experiments.

$$n \approx n' \Leftrightarrow \max_{p \in \{1,\ldots,s\}^3}(|n[p] - n'[p]|) < t_{inc}$$

Pruning and incremental labeling can of course be combined which results in a brick processing scheme as illustrated in Figure 3.

V. EVALUATION AND DISCUSSION

As stated previously, the goal of the proposed framework is to greatly reduce the required runtime and memory footprint while maintaining a high segmentation quality. This is evaluated in this section in the following way: First, the framework is compared to competing methods using the public CT-ORG [65] dataset. Then, as the CT-ORG dataset does not contain datasets exceeding main memory, the proposed method is additionally evaluated using synthetic data which allows for fine grained control over volume sizes and also enables simulation of incremental labeling. Afterwards, the applicability of the method to large real world data is shown in a small case study using datasets from current biomedical research [66] of up to $377\,\text{GB}$ in size (where competing methods are not applicable). Then, we demonstrate using the proposed framework for multi-class segmentation. Finally, we analyze the importance and influence of the brick size parameter $s$ and the introduced pruning methods using synthetic and real data.

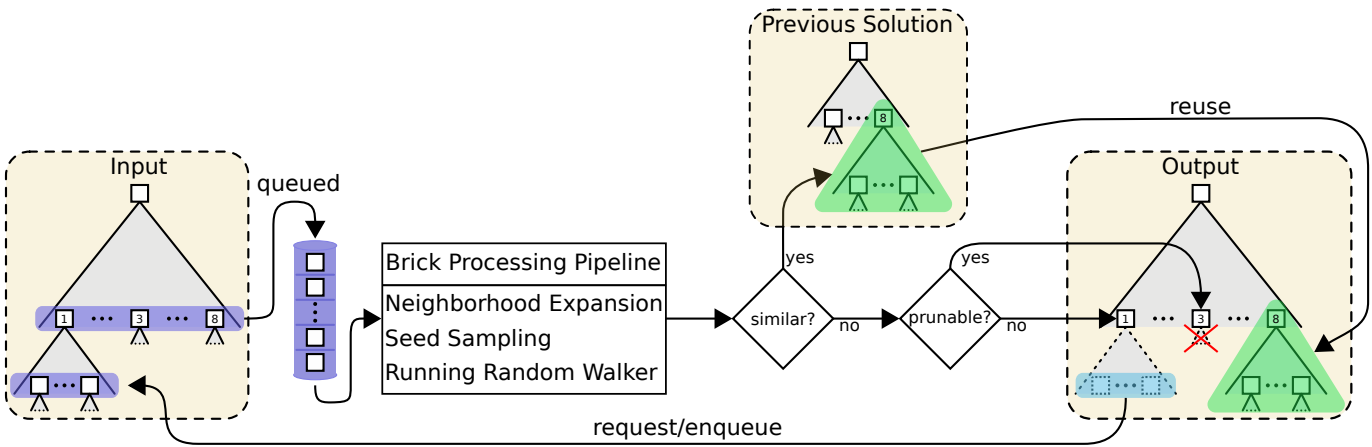

Fig. 3. Illustration of the full brick processing scheme including optimizations: Bricks for a particular level are processed separately (one-by-one, but possibly in parallel) using the core processing pipeline explained in subsection IV-A. Afterwards, bricks are compared to a previous solution and checked for prunability. If the brick is similar to a previous solution (example node 8), the entire subtree can be reused and inserted into the new output tree. If the node can be pruned because it is homogeneous or determined foreground/background (example node 3) the brick is inserted into the output tree, but child nodes are not processed. Otherwise (example node 1), the brick is inserted into the output tree and its 8 children are requested by inserting the corresponding input nodes into the processing queue.

Where applicable, we compare against the standard random walker method [1] (denoted "Basic" in the following), which is also used as a component in the proposed method, and the supervoxel-based method by Fabijanska and Goclawski [50] (denoted "S-Voxel"), which is a recent attempt at reducing the memory footprint and computational burden of the random walker method.

For all methods, the conjugate gradient method [60] with Jacobi preconditioner and an absolute residual magnitude of $10^{-2}$ and 5000 iterations as the stopping criteria was used. Matrices were constructed in the Ellpack format for the proposed method and the original variant, while the CSR format was used for the supervoxel-based method.

All experiments were carried out on a standard PC with an Intel i5-6500 (3.2 GHz, 4 core), 16 GB RAM, a single NVidia GTX 1060 (6 GB VRAM) and a Samsung EVO 970 SSD.

### A. Real World CT Data

For comparison of the proposed framework against the basic random walker and the supervoxel-based approach on real world data, we used the CT-ORG dataset [65], available in TCIA [67], which includes 140 scans in total with associated ground truth segmentation for up to 6 organs types. In order to not exceed the memory capacity of the graphics card when using the basic random walker method, scans were cropped to a $350^3$ voxel region around each organ before evaluation. Scans with less than 350 slices were left out for simplicity, which results in around 80 datasets for each organ except for the brain class, which in general is not visible in most scans in the database. 50 foreground and 200 background seeds were generated by randomly sampling voxel positions from the foreground and background regions. For the supervoxel-based method [50], we chose the same parameters as the authors for CT data: $\beta = 750, V_{MAX} = 5000, \Delta I_{MAX} = 20$. While the t-test-based weight function [29] is used for the proposed framework, the statistical modeling-based weight function [28] was chosen for the basic version since the t-test-based weight function yielded far worse results.

The results are summarized in Table I. Best results (maximum Dice score and minimum time/RAM/VRAM) that differ statistically significantly from the second best result are highlighted in bold. Sample distributions were compared with the Wilcoxon signed-rank test ($\alpha = 0.01$) and Holm-Bonferroni correction for the 24 tests. For all organs, the proposed hierarchical method is clearly faster than the competitor methods: It is faster in comparison with Grady's method by one or two orders of magnitude (2-12 seconds vs. 3-4 minutes) and by at roughly one or more for the supervoxel-based method (2-12 seconds vs. 1-9 minutes). Notably, for the supervoxel method the time variance in general and mean computation time for the brain datasets is very high. This is due to exceptionally high computation time for some datasets – possibly suboptimal parameter choice due to differing imaging conditions.

Similarly for RAM and VRAM requirements the supervoxel method shows an improvement in the mean over the basic method (1.3GB vs. 4.6GB and 0.5GB vs. 5.1GB), but with a high variance and not for the brain dataset. The proposed hierarchical method further reduces both peak RAM and VRAM utilization to roughly 30 and 26MB and does so consistently.

In all but one (brain dataset with only 7 samples) case the superiority of the proposed method in time and memory consumption is shown to be statistically significant.

Looking at the achieved Dice score, it can be seen that the method achieves the goal of maintaining a high segmentation quality when compared to the basic method: For liver, bladder and bone, the proposed variant outperforms the other methods with statistical significance. For lungs, the basic method achieves a better result with statistical significance, but the scores for all methods are within one standard deviation (0.03) of one another. Results for brain and kidneys are not significant. We therefore conclude that the proposed method is actually a reasonable approximation to the basic method and can even achieve substantially better results in some cases.

TABLE I
ACCURACY AND RUN TIME PERFORMANCE COMPARISON OF THE BASIC RANDOM WALKER METHOD [1], SUPERVOXEL-BASED METHOD [50] AND THE PROPOSED HIERARCHICAL VARIANT. WE REPORT THE MEAN AND STANDARD DEVIATION FOR THE TOTAL REQUIRED RUN TIME, PEAK RAM AND VRAM UTILIZATION AND THE DICE SCORE WHEN COMPARED TO THE GROUND TRUTH SEGMENTATION. BEST RESULTS THAT ARE STATISTICALLY SIGNIFICANTLY BETTER THAN THE RUNNER-UPS ARE MARKED IN BOLD. THE PROPOSED OCTREE-BASED METHOD OUTPERFORMS OTHER METHODS IN TERMS OF RUN TIME, RAM AND VRAM REQUIREMENTS WHILE MAINTAINING HIGH SEGMENTATION QUALITY AND OUTPERFORMING COMPETITOR METHODS IN THREE SIGNIFICANT CASES.

|  | Organ | Liver | Bladder | Lungs | Kidneys | Bone | Brain |
|---|---|---|---|---|---|---|---|
|  | #Datasets | 80 | 77 | 81 | 81 | 81 | 7 |
| Dice Score | Basic [1] | 0.788 ± 0.18 | 0.705 ± 0.18 | **0.964 ± 0.03** | 0.908 ± 0.11 | 0.396 ± 0.12 | 0.818 ± 0.03 |
|  | S-Voxel [50] | 0.803 ± 0.11 | 0.511 ± 0.16 | 0.960 ± 0.03 | 0.813 ± 0.12 | 0.583 ± 0.11 | 0.825 ± 0.04 |
|  | Octree (ours) | **0.915 ± 0.04** | **0.770 ± 0.17** | 0.958 ± 0.03 | 0.915 ± 0.06 | **0.738 ± 0.09** | 0.646 ± 0.06 |
| Time [s] | Basic [1] | 207.23 ± 14.54 | 209.07 ± 11.45 | 213.67 ± 12.18 | 223.95 ± 14.58 | 214.41 ± 13.64 | 224.96 ± 14.73 |
|  | S-Voxel [50] | 67.12 ± 138.68 | 70.86 ± 149.88 | 73.52 ± 154.44 | 67.82 ± 140.25 | 73.14 ± 158.30 | 540.72 ± 63.36 |
|  | Octree (ours) | **7.98 ± 2.18** | **2.68 ± 1.17** | **11.68 ± 5.39** | **3.66 ± 0.95** | **8.56 ± 1.35** | **2.09 ± 0.16** |
| RAM [GB] | Basic [1] | 4.60 ± 0.03 | 4.60 ± 0.03 | 4.60 ± 0.03 | 4.60 ± 0.03 | 4.60 ± 0.03 | 4.67 ± 0.00 |
|  | S-Voxel [50] | 1.32 ± 1.78 | 1.34 ± 1.81 | 1.32 ± 1.77 | 1.32 ± 1.77 | 1.30 ± 1.73 | 6.91 ± 0.57 |
|  | Octree (ours) | **0.03 ± 0.00** | **0.03 ± 0.00** | **0.03 ± 0.00** | **0.03 ± 0.00** | **0.03 ± 0.00** | 0.03 ± 0.00 |
| VRAM [GB] | Basic [1] | 5.14 ± 0.00 | 5.14 ± 0.00 | 5.14 ± 0.00 | 5.14 ± 0.00 | 5.14 ± 0.00 | 5.14 ± 0.00 |
|  | S-Voxel [50] | 0.48 ± 1.01 | 0.48 ± 1.03 | 0.52 ± 0.99 | 0.48 ± 1.00 | 0.47 ± 0.97 | 3.63 ± 0.33 |
|  | Octree (ours) | **0.03 ± 0.00** | **0.03 ± 0.00** | **0.02 ± 0.00** | **0.03 ± 0.00** | **0.02 ± 0.00** | 0.02 ± 0.00 |

Having confirmed the expected run time, memory and accuracy characteristics on real world data, in the next section we will expand the evaluation using synthetic data which allows full control over dataset sizes and thus observation of trends in accuracy and – importantly – run time and memory scalability.

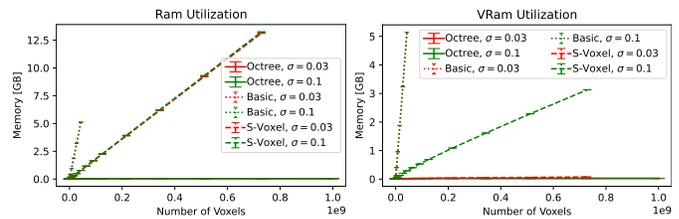

Fig. 4. Median peak main memory (left) and video memory (right) consumption for all examined methods and varying volume sizes. Error bars show minimum/maximum run times (vessel and cell scenarios combined).

### B. Synthetic Data

For systematic comparison with the basic approach on different scales, we generate a ground-truth volume (binary segmentation), a noisy input volume (with a mean of 0.3/0.7 for background/foreground and additive Gaussian noise with varying $\sigma$), as well as foreground and background labels defined as points. Foreground structures cover 1% of the total volume. The *structure size* range is a parameter that influences the size of individual foreground objects and is relative to the dataset size. The upper bound (*structure size* in the following) is 10% if not specified otherwise. The lower bound is always half of the *structure size*. Existing frameworks for biomedical test data generation [68] are limited in the size of datasets which can be generated. Therefore we define structures as described below and create a voxel representation by organizing the primitives (spheres, cylinders) in sorted queues (according to their bounding boxes) for the $z$, $y$ and $x$-dimension, scanning the output volume linearly, for each voxel rasterizing active primitives [69]. We chose the following biomedical application-inspired scenarios:

*Cells* are represented as spheres with a diameter within structure size and placed randomly within the volume. Centers of spheres do not intersect other spheres, thus reducing overlap, but allowing for "clustering" of cells. Foreground seeds are placed at the center of each sphere. For each generated sphere 2 background seed points are placed at a random position with a distance of two radii from the cell center, but outside of any other cell.

*Vessels* consist of segments of cylindrical shape and spheres at branching points. Each vessel tree starts with a radial diameter within the structure size from a random point at the volume border. The vessel branches at $5 \times$ radius into two segments, preserving cross-section area. This process repeats recursively. Two points along the vessel centerline are added to the set of foreground labels while two corresponding points outside the vessel with a distance of twice the radius are added as background labels.

All experiments have been executed 5 times with different seeds for the random number generator so that minimum, median and maximum performance can be evaluated in the following. For the supervoxel-based method [50], we empirically chose the following parameters, which were subjectively optimal on a selected number of exemplary datasets: $\beta = 8, V_{MAX} = \frac{d_x d_y d_z}{30^3}, \Delta I_{MAX} = \frac{1}{20}$. It should be noted, however, that especially for $\Delta I_{MAX}$ there is a trade-off between accuracy and memory/run time performance. For both the proposed framework and the basic variant, the t-test-based weight function [29] is used. We first show our main contributions (reduction of memory consumption enabling out-of-core application, run time improvements) before demonstrating competitive accuracy of the method.

*1) Memory Consumption:* Figure 4 shows the peak main memory (RAM) and video memory (VRAM) consumption of the compared methods for increasing volume sizes up to $1000^3$ voxels. Notably, for the basic random walker method, both RAM and VRAM consumption rise sharply such that for volumes of $400^3$ voxels the method failed due to insufficient available VRAM (more than 6GB).

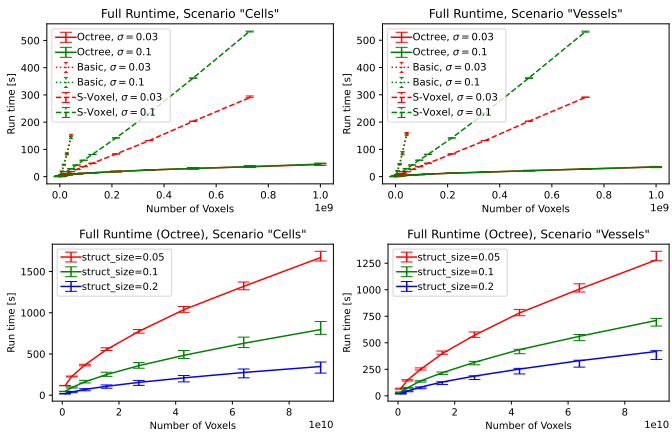
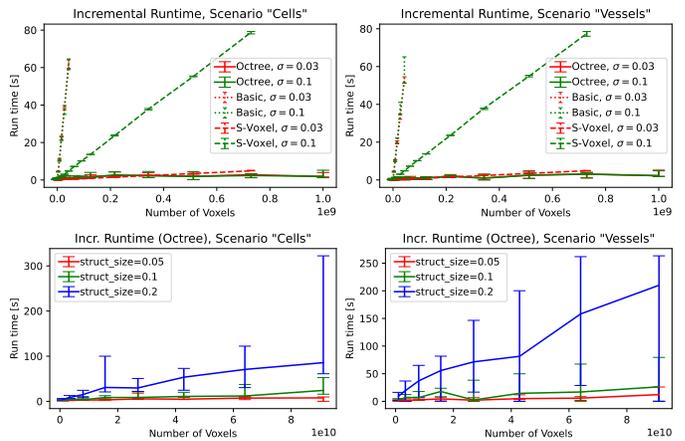

Fig. 5. Median wall-clock run time of examined methods (top row), and octree variant for larger volume sizes and varying structure sizes (bottom row). Error bars show minimum/maximum run times.

Fig. 6. Median wall-clock run time after adding a label for a previously unlabeled foreground component for examined methods (top row), and for the proposed octree variant with larger volume sizes and varying structure sizes (bottom row). Error bars show minimum/maximum run times.

For the supervoxel-based method, its main idea of reducing the size of the equation system to be solved is clearly visible in the VRAM plot. It should be noted, however, that the success is highly dependent on imaging conditions: For $\sigma = 0.03$ the system size is indeed only a few megabytes, while for $\sigma = 0.1$ a large number of single voxels as supervoxels are produced, resulting consumption of more than 3GB for volumes of $900^3$ voxels. Further, the required *main memory* is still linearly dependent on the size of the input image since the image itself and auxiliary data structures have to be loaded into RAM during the supervoxel-construction phase. This ultimately means that the method cannot process volumes larger than $900^3$ in this experiment, where the memory requirements already exceed 13GB independent of the noise present in the image.

In contrast to this, due to iterative loading and processing of constant sized bricks (see Figure 3) the proposed octree-based method requires less than 450 MB of RAM and less than 30MB of VRAM even for volumes of $4500^3$ voxels ($182\,\mathrm{GB}$ in size). This demonstrates one of our main contributions, i.e., the ability to process larger-than-main memory volumes.

*2) Full Run Time:* Figure 5 (top row) compares the run time of the examined methods. The bottom row only shows the octree variant, up to a volume size of $4500 \times 4500 \times 4500$ voxels ($182\,\mathrm{GB}$). In direct comparison, the octree variant is considerably faster than the supervoxel-based method, which in turn requires less computation time than the basic random walker method. This becomes even more apparent for increasing dataset sizes. For datasets of $350^3$ voxels the original variant requires more than two minutes of computation time (independent of the scenario) while supervoxel method requires roughly 30 seconds and the proposed method completes in less than $9\,\mathrm{s}$ in all cases. For the supervoxel method, the run time is also dependent on the level of noise in the image: A volume of $900^3$ voxels requires either roughly 5 ($\sigma = 0.03$) or 9 minutes ($\sigma = 0.1$) to be processed (due to differences in the size of the linear equation system to be solved). The proposed method in contrast finishes in less than $40\,\mathrm{s}$. Furthermore, while the run time for both the basic and supervoxel-based approach is shown to be linear, the run time of the proposed framework appears sub-linear (see Figure 5, bottom row). This is consistent with the runtime behavior predicted in subsubsection IV-D1: Due to pruning, at any level only bricks intersecting a foreground-background boundary within the volume require further processing, ultimately resulting in a runtime of $v^{\frac{2}{3}}$ for a volume of $v$ voxels. This hypothesis is further supported by the fact that smaller structure sizes – resulting in a higher *number* of structures and ultimately more surface area – cause longer computation times.

*3) Run time with Simulation of Incremental Labeling:* In order to demonstrate run time savings of the incremental labeling aspect of the proposed approach, two subsequent runs are performed to simulate a single user editing step. In the first run, a randomly selected foreground label (a single center point of a sphere or a vessel segment) is retained from the complete set of labels. For the second run this label is added and the run time is measured.

Figure 6 shows the incremental run time for the basic, the supervoxel-based and the proposed octree random walker approach in comparison (top row), and for the octree variant for datasets larger than $1000^3$ voxels (bottom row). Overall, all methods appear to benefit from the additional information of a previous solution. Specifically, for the basic random walker variant, at $350^3$ voxel the average run time is reduced by more than half to roughly to 60 seconds. For the supervoxel-based method, reusing a previous solution involves skipping the expensive preprocessing step, and also reusing the previous solution as an initialization for solving the (unchanged) linear equation system. This also shows in a major reduction of run time which, however, depends on the noise in the image: For noisier images, the linear equation system is considerably larger due to the larger number of supervoxels found, which results in a run time of more than one minute ($\sigma = 0.1$) and roughly 5 seconds ($\sigma = 0.03$) for a volume of $900^3$ voxels.

The proposed hierarchical variant shows moderate increase in computation time from less than 200 milliseconds for the smallest dataset ($50^3$ voxels) to consistently less than 4 seconds for the $900^3$ voxel dataset, which is a further reduction in comparison with the full (non-incremental run time) shown in Figure 5.

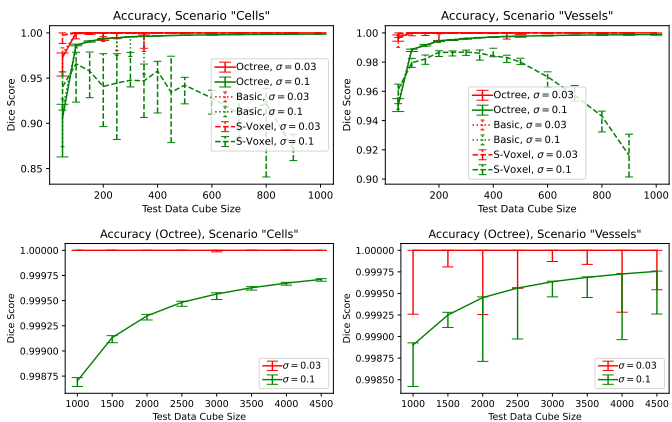

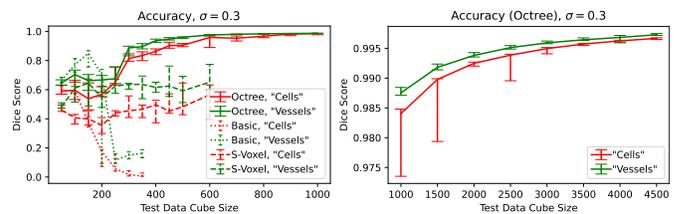

Fig. 7. Median Dice score of the generated segmentation compared to ground truth for examined methods (top row), and only the proposed octree variant for larger volume sizes (bottom row). Error bars show minimum/maximum scores.

Fig. 8. Median Dice score of the generated segmentations compared to ground truth for high noise levels ($\sigma = 0.3$). Error bars show minimum/maximum scores.

In contrast to the full run time evaluation, structure sizes have a different effect: The overall trend is that larger structure sizes require more computation time. This is expected, as (in contrast to the full computation) only a single structure element requires modification. Smaller structures thus require less work overall. For a structure size of 0.05, the hierarchical method completes in a median time of 10 seconds, even for datasets of $4500^3$ voxels (182 GB). Larger structures (structure size 0.2, spanning large portion of volume) result in higher processing times up to roughly 85 seconds (cells) and 210 seconds (vessels) in the median and 320 seconds (cells) and 260 seconds (vessels) in the maximum. This is significantly lower than for the full computation and still usable in practice considering that volumes with large structures likely require few labels overall.

*4) Accuracy:* To evaluate the accuracy of the examined methods, the generated segmentation (using a threshold of 0.5 on the foreground probability map) was compared to the ground truth segmentation using the Dice score. First, we review results for low ($\sigma = 0.03$) and medium ($\sigma = 0.1$) noise levels. For $\sigma = 0.03$ the resulting overlapping coefficient [70] between foreground and background intensity distributions (i.e., the overlapping area of the density functions) equals roughly $10^{-11}$. This means that the foreground and background intensity distributions are largely separate. For $\sigma = 0.1$ the distributions overlap moderately (overlapping coefficient 0.046).

Figure 7 (top row) compares the accuracy of the basic random walker, the supervoxel-based approach and the proposed octree variant on synthetic datasets up to a volume size of $1000^3$ voxels. The bottom row again only the octree variant, up to a volume size of $4500^3$ voxels.

For low noise levels ($\sigma = 0.03$) all methods achieve high medium Dice scores, and actually yield a perfect score of 1 in the majority of cases. For medium noise levels ($\sigma = 0.1$) the basic and the proposed variant achieve similar scores with the accuracy of the proposed variant improving for larger volumes. This trend also continues for volumes up to $4500^3$ voxels where the minimum Dice score is above 0.99925 in both scenarios. The performance of the supervoxel-based method, however, drops for medium noise levels (in some cases below a Dice score of 0.9). The reason for this is that the supervoxel construction method fails to detect the border of structures in some cases which results in supervoxels that lie in both foreground and background. This error cannot be fixed in the later stage of the pipeline.

As noted above the differences between both noise levels appear to reduce as the volume resolution is increased. This motivates the extension of the above comparison to high noise levels ($\sigma = 0.3$) where foreground and background intensity distributions overlap to a large degree (overlapping coefficient [70] of 0.50). Dice scores for small volumes and both variants (left) as well as for large volumes and the hierarchical method (right) are shown in Figure 8. It is apparent that all variants struggle to produce the expected result for small volume sizes due to the extreme noise up to a size of roughly $250^3$ voxels. This is especially true for the cell scenario – possibly due to fewer seeds points from the dataset generation. Here, the behavior of three methods diverges: The resulting score for the basic random walker methods sharply drops to below 0.2. The supervoxel-based method continues to achieve Dice scores between 0.4 and 0.8. Notably, the maximum volume size that could be processed also reduced to $600^3$ due to the increase in generated supervoxels and the resulting larger linear equation system size and ram consumption. However, the accuracy of the proposed hierarchical variant once again starts to improve: For volumes of $1000^3$ voxels, Dice scores consistently above 0.97 are achieved and for $4500^3$ Dice scores are above 0.995. The mean filtering of the LOD approach can be expected to reduce the (additive) noise at the coarser levels of the pyramid, allowing the broad structure of foreground objects to be captured, even when the exact surface features cannot be reconstructed due to the noise.

The Dice scores were observed to be equivalent between incremental and non-incremental runs and are thus not discussed further here. Similarly, the data structure size was not observed to have a noticeable impact on the accuracy of the method.

## C. Small Case Study: Large Real Data

In the previous two subsections we have shown that the proposed hierarchical variant of the random walker method is comparable to the basic method in accuracy and that it has substantial benefit in terms of the required computational time and memory requirements.

Here, as part of a small case study, we additionally apply the proposed method to real multi-gigabyte datasets far exceeding the limits of available main memory that are used in current biomedical research [66] where the other considered methods are not applicable due to the massive data size. Comparison with ground truth data can not be performed in this case since ground truth information is not available – at least in part due to the lack of labeling tools capable of handling data of this size, a gap that our work aims to fill. The processed datasets include an antibody-stained light sheet microscopy scan of a mouse kidney and a scan of a mouse brain hemisphere[3]. In total 4 labeling tasks were carried out: 1) A binary segmentation of the arterial vasculature (up to the capillary level) created on a $60\,\text{GB}$ subset of the whole kidney dataset; 2) The vascular tree of the full $377\,\text{GB}$ kidney dataset up to a vessel diameter of about $10\,\mu\text{m}$ and 3) a foreground mask for the whole kidney volume were created in the same way; 4) The vasculature of the $59\,\text{GB}$ brain dataset up to a vessel diameter of roughly $10\,\mu\text{m}$ was created similarly. Figure 9 shows a rendering of the original datasets, in conjunction with the vasculature and whole kidney segmentation for illustration (left), a subset of the kidney dataset with a more detailed segmentation of the vasculature (center) and the brain dataset with the vasculature segmentation (right).

Table II shows some additional information about the datasets, including the full computation time (without a previous result, given the full set of labels) and observed computation time ranges for updates to the label set. The observed full computation time is in line with the results of the synthetic dataset experiments (see Figure 5) and thus substantial, but satisfactory in practice (below one hour) given the size of the datasets. For vasculature the incremental computation times are observed to be in the order of seconds (especially for smaller vessels (1s), but also for the long sections of the largest vessels (60s)) and are thus dominated or roughly equivalent to the time a user needs for inspection and interaction with the data. For the full kidney segmentation individual changes to the label set require a longer run time to update the output probability map (between few minutes and the full run time) in this experiment. This, however, is balanced out by the fact that very few labels are required overall. In general, regions of high uncertainty (i.e., foreground probability values around 0.5) cause longer computation times since those regions often cannot be pruned. This was also the case for the kidney, where three foreground labels in the center and two on the outside were able to cleanly separate foreground from background

---

[3]All procedures involving animals (i.e., procedures performed to obtain these datasets) were carried out in strict accordance with the German animal protection legislation (Tierschutzgesetz und Tierschutzversuchstierverordnung). The protocol was approved (84-02.04.2016.A218) by the Committee on the Ethics of Animal Experiments of the Landesamt für Natur, Umwelt und Verbraucherschutz (LANUV).

---

at the capsule, but failed to define a clear boundary at the hilum where vessels are entering the kidney. Here, a number of additional labels were required to remove the ambiguity. Thus, a sudden increase of computation time signifies resulting uncertainty in the segmentation and may prompt to correct the last set of labels and restart the computation.

The total label time (including user interaction) was roughly three hours for the large vasculature in the full $377\,\text{GB}$ dataset, and about seven hours for the $60\,\text{GB}$ subset as well as the $59\,\text{GB}$ brain dataset, both with substantially more detail. The broad majority of this time can be attributed to the user interaction and identification of vessels itself, which is especially challenging for the kidney datasets since segmented vessels are often of comparatively low intensity, which also makes the application of fully automated method difficult. The segmentation of the whole kidney is dominated by the full compute time since the initial set of labels already generated a good segmentation of the whole kidney and few corrections were required. The user interaction time itself was in the order of minutes.

## D. Demonstration: Multi-Class Segmentation

In order to demonstrate multi-class segmentation, the proposed method was applied to a volume form the CT-ORG dataset [65]. Seeds were placed in lungs, liver, kidneys and background and the method run four times to obtain probability maps for each of the classes, and the output class is obtained using a maximum operation as described in section IV. Since in this non-binary segmentation scenario the assumption for $\text{dt}$-based pruning does not hold (see subsubsection IV-D1), only homogeneity-based pruning with $t_{hom} = 0.3$ was applied. The generated class maps in conjunction with the original volume are visualized in Figure 10.

It should further be noted that multi-class segmentation is another use case where pruning and reusing of parts of the old solution prove beneficial: In the above example, when a user place an initial seed of the *kidney* class into the previously unlabeled organ, for the *lung* and *liver* probability maps, these appear as additional background labels far from any foreground labels which thus influence the solution very little. The following recomputations thus finish in fractions of a second for these classes.

## E. Influence of Brick Size s

In the initial presentation of the algorithm and during all previous experiments, the brick size $s$ was set to the fixed value of 32. In this section we justify this selection using experiments on the previously discussed synthetic datasets of increasing sizes. The results are presented in Figure 11 including run time and accuracy in terms of the Dice score.

As demonstrated in the top row, an increase in $s$ also results in an increased run time of the algorithm. Specifically, a brick size of $s = 64$ requires more than double the run time compared to $s = 32$, but $s = 16$ only offers a small advantage over $s = 32$. This is likely due to larger brick sizes causing the method to benefit less from optimizations such as pruning: The bulk of work (roughly $\frac{3}{4}$ since only the object surface needs to be refined, see section A) is spent processing the leaf layer of the LOD pyramid where a finer subdivision into bricks (due to smaller $s$) causes fewer voxels to be processed overall due to finer-grained pruning. Processing additional layers in a higher LOD pyramid (due to the smaller $s$) is insignificant in comparison.

TABLE II
INFORMATION ABOUT THE PROCESSED REAL WORLD DATASETS INCLUDING FULL RUN TIME (WITHOUT PREVIOUS RESULTS AND WITH THE FULL LABEL SET) AND TIME RANGES FOR LABEL UPDATE OPERATIONS.

| Dataset | Voxel Dimensions | Real Dimensions | Size | Full Run Time | Incr. Run Time |
|---|---|---|---|---|---|
| Kidney Vasculature (Subset) | $5153 \times 4791 \times 1213$ | $3.9 \times 3.6 \times 3.6 \, \text{mm}^3$ | 60 GB | 10 min | 1 s–15 s |
| Kidney Vasculature (Full) | $9070 \times 12732 \times 1634$ | $6.8 \times 9.6 \times 4.9 \, \text{mm}^3$ | 377 GB | 11 min | 1 s–60 s |
| Full Kidney (Full) | $9070 \times 12732 \times 1634$ | $6.8 \times 9.6 \times 4.9 \, \text{mm}^3$ | 377 GB | 48 min | 4 min–48 min |
| Brain Vasculature | $3895 \times 4215 \times 1801$ | $6.3 \times 6.8 \times 3.6 \, \text{mm}^3$ | 59 GB | 7 min | 1 s–17 s |

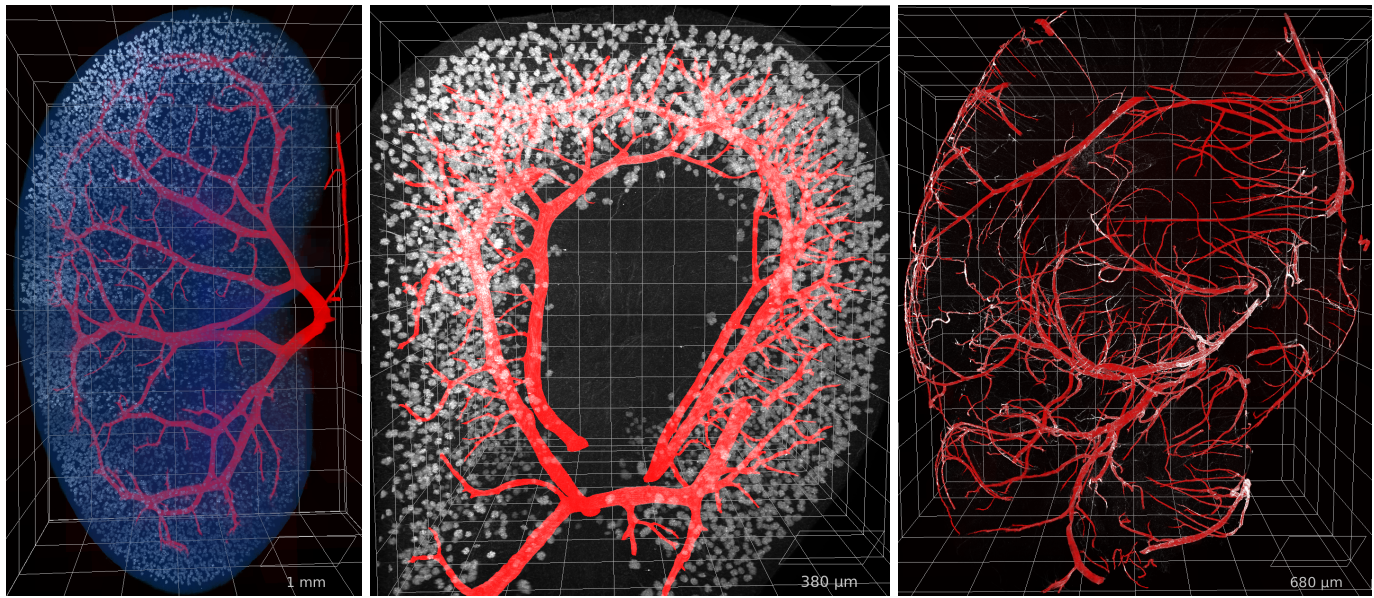

Fig. 9. Ray casting rendering of the kidney dataset (377 GB left, 60 GB center) mainly showing high intensity glomeruli (grayscale) and of the brain dataset (59 GB, right), including the vessels tree (red) and kidney volume (blue) segmentation obtained using the proposed method. The resulting vessel networks have a total length of 164 mm (left), 170 mm (center), 428 mm (right) and a volume of $0.67 \, \text{mm}^3$ (left), $0.35 \, \text{mm}^3$ (center) and $0.32 \, \text{mm}^3$ (right).

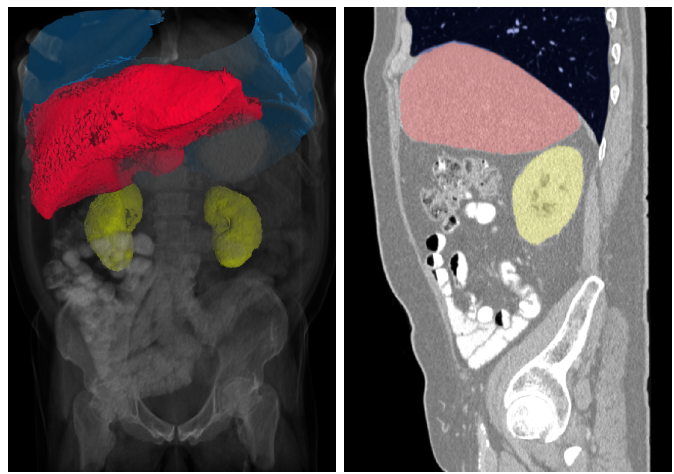

Fig. 10. Demonstration of multi-class segmentation using the proposed method on a volume from the CT-ORG dataset [65]. The four segmented classes (background (transparent), lung (blue), liver (red) and kidneys (yellow)) are shown with the original dataset in a 2D slice (right) and using direct volume rendering (left).

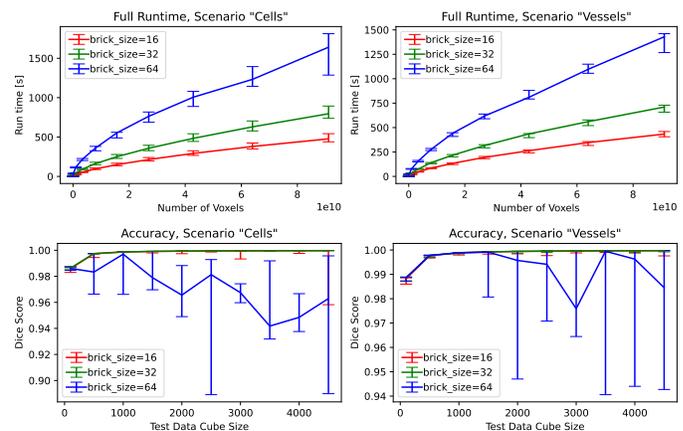

Fig. 11. Median wall-clock run time (top row) and Dice score (bottom row) of the presented method for varying brick sizes $s$. Error bars show minimum/maximum values.

On the other hand, a small brick size of $s = 16$ reduces the accuracy of the method significantly in some cases (with Dice scores below 0.96 in one case). Notably, choosing a larger brick size of $s = 64$ also results in reduced Dice scores of below than 0.9 in some cases which is likely due to the "strictness" of the selected weight function [29] which sometimes fails to propagate labels far from unfavorable seeds positions, which is mitigated by using smaller brick sizes processing otherwise. Notably, with the statistical modeling-based weight function [28] accuracy *increases* for a higher brick size (not shown here). Finally, it is easy to derive that a doubling of $s$ results in an 8-fold increase in memory requirements, which, however, as shown in subsection V-B, is entirely unproblematic even for $s = 64$. For these reasons we suggest setting $s = 32$ as a sweet-spot in the trade-off between accuracy and run-time *independent of image resolution.*

TABLE III
MEAN DICE SCORE AND RUNTIME ON THE CT-ORG REAL WORLD DATA SETS FOR DIFFERENT COMBINATIONS OF PRUNING METHODS: HOMOGENEITY-BASED PRUNING (ACCORDING TO hom) AND PRUNING OF ALREADY DETERMINED BRANCHES (ACCORDING TO dt).

| hom-Pruning | dt-Pruning | Runtime | Dice Score |
|---|---|---|---|
| No | No | 32.15s | 0.85616 |
| Yes | No | 8.00s | 0.85617 |
| No | Yes | 6.87s | 0.85610 |
| Yes | Yes | 6.86s | 0.85611 |

TABLE IV
MEAN DICE SCORE AS WELL AS FULL AND INCREMENTAL RUNTIME (RT) ON 40 SYNTHETIC DATASETS FOR DIFFERENT COMBINATIONS OF PRUNING METHODS: HOMOGENEITY-BASED PRUNING (ACCORDING TO hom) AND PRUNING OF ALREADY DETERMINED BRANCHES (ACCORDING TO dt).

| hom-Pruning | dt-Pruning | Full RT | Incr. RT | Dice Sc. |
|---|---|---|---|---|
| No | No | 767.94s | 5.21s | 0.99877 |
| Yes | No | 38.57s | 2.47s | 0.99877 |
| No | Yes | 38.38s | 2.46s | 0.99877 |
| Yes | Yes | 38.42s | 2.46s | 0.99877 |

*F. Influence of Pruning Methods*

Apart from the reduced memory footprint, the hierarchical processing scheme introduced in this paper may not by itself bring large advantages in terms of total runtime. This is achieved by pruning branches of the probability map octree during its construction based on the criteria hom (homogeneity-based) and dt (pruning of already determined complete foreground or background regions). In this section we will determine the influence of both methods by observing results when disabling one of two or both pruning methods. For both pruning methods the thresholds suggested in section IV were applied ($t_{hom} = 0.3$ and $t_{bin} = 0.01$).

Table III shows the mean runtime and Dice score when applying the method to the same CT datasets as in subsection V-C. The pruning methods appear to have a negligible influence on the segmentation accuracy: The difference in Dice score is insignificant (less than 0.1%) between any of the examined configurations. Both pruning methods reduce the overall runtime when applied on their own: hom-based pruning reduces the full runtime to roughly a quarter in comparison to no pruning, while dt-based reduces the time by another 15%. Combining both pruning methods yields the same speedup as only applying dt-based pruning.

A similar evaluation was also conducted for a total of 40 artificial datasets, constructed as described in subsection V-B with 20 datasets from both scenarios. All datasets consist of $1000^3$ voxels with a structure size of 10%. The results in Table IV are overall similar to the results on the CT dataset: No difference in accuracy between pruning methods (or no pruning method) could be detected. Further, both pruning methods exhibit a substantial speedup for a full run by a factor of roughly 20 when compared to running without pruning. Here, there is very little difference in speedup between both pruning methods on their own . Once again, combining both pruning methods does not result in an additional speedup. It should be noted however, that – as described in section IV – dt-based pruning only makes sense if the desired end result is a binary segmentation. If on the other hand the foreground probability map (for example for uncertainty visualization) is desired or in case of multi-class segmentation, it may be useful to only apply homogeneity-based pruning which still achieves a considerable speedup on synthetic and real data.

Table IV also shows that the reusing of branches for incremental labeling (when adding a single foreground label in this experiment) results in a substantial speedup by a factor of roughly 20 even when not pruning branches. At the same time, applying any of the two or both pruning methods further halves the incremental runtime.

VI. CONCLUSION

We have presented a scalable approximation of the random walker method for semi-automatic segmentation of very large volumetric datasets. In contrast to the related hierarchical graph-cuts method [44] our method can be used for multi-class segmentation, can reuse previous solutions for an additional speedup and – due to the brick-wise processing – can be applied to volumes that exceed the main and graphics memory capacity. Our method outperforms competitor variants[1], [50] in run time performance while preserving high segmentation accuracy on both artificial and real test sets. While other considered approaches fail to process datasets larger than some hundreds of megabytes due to video or main memory exhaustion, here the applicability to large real world datasets of up to 377 GB has been demonstrated in a case study which also confirmed the runtime behavior observed on controlled synthetic test data. The approach can process large datasets on consumer PCs, but at the same time is future-proof and scalable beyond the experiments in this paper, as it is easily parallelizable and should be able to fully utilize multi-GPU setups. As such, it should be useful, for example, as a stage in a pipeline for quantitative analysis of large volumetric datasets, or as a highly efficient labeling tool for machine learning applications. An implementation of the presented method has been published as part of version 5.2 of Voreen [71], a widely used open source volume rendering and processing framework.

In the future, we will explore the integration of extensions to random walker algorithm [9] such as automatic label generation [26]. We will further improve the current implementation by moving more computations, such as preprocessing and linear equation system construction, from the CPU onto the GPU, thus further improving the scalability. Furthermore, it is worth investigating the use of other compatible methods that compute segmentation probability maps from user-defined seeds in our framework that is of general nature.

## ACKNOWLEDGEMENT

This work was funded by the Deutsche Forschungsgemeinschaft (DFG) – CRC 1450 – 431460824.

## APPENDIX

### A PROOF FOR $\mathcal{O}(B^{\frac{2}{3}})$ GROWTH OF THE NUMBER OF BRICKS INTERSECTING THE OBJECT SURFACE

**Definitions:** Let $P = [0, N] \times [0, M]$ be a parameter space and let $V = [0,1]^3 \subset \mathbb{R}^3$ define a cubic volume in 3D. $\phi : P \to V$ defines the surface of a foreground object in $V$ using the parameter space $P$. We restrict $\phi$ to be able to "squish" the surface, but not "stretch" it, i.e.,

$$\forall x, y \in P : ||x - y||_2 \geq ||\phi(x) - \phi(y)||_A \quad (5)$$

where $||\cdot||_A$ denotes the distance on the surface $A$. In practice this is not a limitation, since $P$ can be chosen as large as required in order for $\phi$ to only "squish".

For $V$ we define a grid-like subdivision into $n^3 =: B$ cubic bricks $C_{klm}$ (i.e., $L_\infty$ balls) of side length $d \times d \times d$, with $d = \frac{1}{n}$ and $(k, l, m) \in \{1, ..., n\}^3 =: I$. These bricks $C_{klm}$ completely cover V, i.e.: $\bigcup_{s \in I} C_s = V$. In particular this means that the centers $c_{klm}$ of bricks $C_{klm}$ have the following $L_\infty$-distance relationship:

$$\forall (k, l, m), (k', l', m') \in I : ||c_{k,l,m} - c_{k',l',m'}||_\infty$$
$$= \max(|d(k - k')|, |d(l - l')|, |d(m - m')|)$$
$$\stackrel{d \geq 0}{=} d \max(|k - k'|, |l - l'|, |m - m'|) \quad (6)$$

We can count the number of bricks that intersect with the surface $A$ with the following function $b(A)$:

$$b(A) = |\{r \in I \mid C_r \cap A \neq \emptyset\}|$$

**Statement:** The number of bricks that intersect the object surface $A$ grows with the $\frac{2}{3}$ of $B$, i.e.: $b(A) \in \mathcal{O}(B^{\frac{2}{3}})$.

**Proof:** The euclidean distance between two points on the surface $A = \phi(P)$ is bounded by the surface distance, i.e.:

$$\forall x, y \in P : ||\phi(x) - \phi(y)||_A \geq ||\phi(x) - \phi(y)||_2$$
$$\stackrel{\text{Equation 5}}{\Rightarrow} ||x - y||_2 \geq ||\phi(x) - \phi(y)||_2 \quad (7)$$

Similar to the grid-subdivision of $V$, we divide the parameter space $P$ up into $p \cdot q$ squares, where $p = \frac{N}{d} = Nn$, $q = \frac{M}{d} = Mn$ and squares $S_{ij}$ (i.e. $L_\infty$-balls) of size $d \times d$ with $i \in \{1, ..., p\}$, $j \in \{1, ..., q\}$. Without loss of generality let $N, M \in \mathbb{N}$ and thus also $p, q \in \mathbb{N}$. If this is not the case, we can simply replace $\phi_{new} = c \circ \phi$ with an element-wise scaling function $c$. Let $s_{ij}$ be the center of the $L_\infty$-ball $S_{ij}$. Then, a point from $S_{ij}$ has euclidean distance of less than $d$ to $s_{ij}$ after application of $\phi$:

$$\forall x \in S_{ij} : ||\phi(x) - \phi(s_{ij})||_2 \stackrel{\text{Equation 7}}{\leq} ||x - s_{ij}||_2$$
$$\stackrel{dim(P)=2}{\leq} \sqrt{2}||x - s_{ij}||_\infty \stackrel{x \in S_{ij}}{\leq} \sqrt{2}\frac{d}{2} = \frac{d}{\sqrt{2}} < d \quad (8)$$

Now, let $x \in S_{ij}$ and $\phi(s_{ij}) \in C_u$ and $\phi(x) \in C_t$. Then, using the triangle inequality, we can find a strict upper bound for the distance between the centers of the bricks of $\phi(x)$ and $\phi(s_{ij})$:

$$||c_u - c_t||_\infty$$
$$\leq ||c_u - \phi(s_{ij})||_\infty + ||\phi(s_{ij}) - \phi(x)||_\infty + ||\phi(x) - c_t||_\infty$$
$$\stackrel{\phi(s_{ij}) \in C_u \wedge \phi(x) \in C_t}{\leq} \frac{d}{2} + ||\phi(x) - \phi(y)||_\infty + \frac{d}{2}$$
$$\leq \frac{d}{2} + ||\phi(x) - \phi(y)||_2 + \frac{d}{2} \stackrel{\text{Equation 8}}{<} \frac{d}{2} + d + \frac{d}{2} = 2d \quad (9)$$

Hence, we know that the index difference $u - t = (k', l', m') \in \mathbb{Z}$ of the brick of $\phi(s_{ij})$ ($C_u$) and the one that contains $\phi(x)$ ($C_t$) is restricted:

$$d \max(|k'|, |l'|, |m'|) \stackrel{\text{Equation 6}}{\leq} ||c_u - c_t||_\infty \stackrel{\text{Equation 9}}{<} 2d$$
$$\Rightarrow |k'| < 2 \wedge |l'| < 2 \wedge |m'| < 2$$
$$\stackrel{k',l',m' \in \mathbb{Z}}{\Rightarrow} k', l', m' \in \{-1, 0, 1\} \quad (10)$$

Let $I_u = \bigcup_{(k',l',m') \in \{-1,0,1\}} u + (k', l', m')$ be set of indices of bricks that intersect with the patch $\phi(S_{ij})$. Then we can show that the surface patch $\phi(S_{ij})$ is contained in a (connected) region of at most $|\{-1, 0, 1\}|^3$ bricks:

$$b(\phi(S_{ij})) = |\{r \in I \mid C_r \cap \phi(S_{ij}) \neq \emptyset\}|$$
$$= |\{r \in I_u \mid C_r \cap \phi(S_{ij}) \neq \emptyset\}$$
$$\cup \{r \in I \setminus I_u \mid C_u \cap \phi(S_{ij}) \neq \emptyset\}|$$
$$\stackrel{\text{Equation 10}}{=} |\{r \in I_u \mid C_r \cap \phi(S_{ij}) \neq \emptyset\}|$$
$$\leq |\{I_u\}| = 27 \quad (11)$$

With this information, we can find an upper bound for $b(A)$

$$b(A) = |\{r \in I \mid C_u \cap \bigcup_{ij} \phi(S_{ij}) \neq \emptyset\}|$$
$$= |\{r \in I \mid \bigcup_{ij}(C_u \cap \phi(S_{ij})) \neq \emptyset\}|$$
$$= |\{r \in I \mid \bigvee_{ij}(C_u \cap \phi(S_{ij}) \neq \emptyset)\}|$$
$$= |\bigcup_{ij}\{r \in I \mid C_u \cap \phi(S_{ij}) \neq \emptyset\}|$$
$$\leq \sum_{ij} |\{r \in I \mid C_u \cap \phi(S_{ij}) \neq \emptyset\}|$$
$$= \sum_{ij} b(\phi(S_{ij})) \stackrel{\text{Equation 11}}{\leq} \sum_{ij} 27$$
$$= \sum_{i \in \{1,...,p\}} \sum_{j \in \{1,...,q\}} 27 = 27pq$$
$$= 27NMn^2 = 27NMB^{\frac{2}{3}} \in \mathcal{O}(B^{\frac{2}{3}})$$

since N and M are constants for a given parametrization $\phi$ and surface $A = \phi(P)$. ∎

If the surface of the object is not a single connected component, then the proof above applies to all components individually. Then the total number of bricks is smaller than or equal than the sum of all components, which is of course again within $\mathcal{O}(B^{\frac{2}{3}})$.

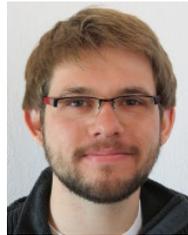

**Dominik Drees** received the M.S. degree in computer science from University of Münster, Germany, in 2016 and is currently pursuing the Ph.D. degree. His research interest are biomedical image processing and analysis with a focus on scalable out-of-core methods.

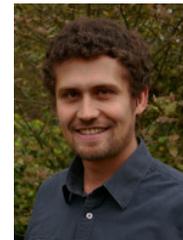

**Florian Eilers** received the M.S. degree in computational life science from University of Lübeck, Germany, in 2020 and is currently pursuing the Ph.D. degree. His research interest are biomedical image processing and analysis with a focus on complex valued images and complex valued neural networks.

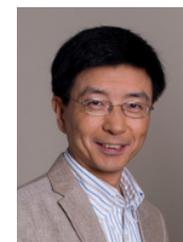

**Xiaoyi Jiang** studied Computer Science at Peking University, China, and received his Ph.D. and Venia Docendi (Habilitation) degree from University of Bern, Switzerland. He was an Associate Professor at Technical University of Berlin, Germany. Since 2002 he is a Full Professor at University of Münster, Germany, where he is currently the Dean of the Faculty of Mathematics and Computer Science. He is an Editor-in-Chief of International Journal of Pattern Recognition and Artificial Intelligence. He also serves on the Advisory Board and Editorial Board of several journals, including International Journal of Neural Systems and Journal of Big Data. Previously, he has been associate editor for IEEE Trans. on Systems, Man, and Cybernetics - Part B / IEEE Trans. on Cybernetics, IEEE Trans. on Medical Imaging, and Pattern Recognition. He is chair of IEEE EMBS Technical Committee on Biomedical Imaging and Image Processing (BIIP). He is a Senior Member of IEEE and Fellow of IAPR.